\documentclass[]{interact}

\usepackage{graphicx}
\usepackage{subfigure}

\usepackage{natbib}
\bibpunct[, ]{(}{)}{;}{a}{}{,}

\begin{document}

\articletype{ARTICLE}

\title{Land Classification in Satellite Images by Injecting Traditional Features to CNN Models}

\author{
\name{Mehmet Çağrı Aksoy\textsuperscript{a} Beril Sirmacek\textsuperscript{b} and Cem Ünsalan\textsuperscript{a}\thanks{CONTACT Cem Ünsalan. Email: cem.unsalan@marmara.edu.tr}}
\affil{\textsuperscript{a}Marmara University, Faculty of Engineering, Department of Electrical and Electronics Eng., Istanbul, Turkey; \textsuperscript{b}Independent scientist, Overijssel, The Netherlands}
}

\maketitle

\begin{abstract}
Deep learning methods have been successfully applied to remote sensing problems for several years. Among these methods, CNN based models have high accuracy in solving the land classification problem using satellite or aerial images. Although these models have high accuracy, this generally comes with large memory size requirements. On the other hand, it is desirable to have small-sized models for applications, such as the ones implemented on unmanned aerial vehicles, with low memory space. Unfortunately, small-sized CNN models do not provide high accuracy as with their large-sized versions. In this study, we propose a novel method to improve the accuracy of CNN models, especially the ones with small size, by injecting traditional features to them. To test the effectiveness of the proposed method, we applied it to the CNN models SqueezeNet, MobileNetV2, ShuffleNetV2, VGG16, and ResNet50V2 having size 0.5 MB to 528 MB. We used the sample mean, gray level co-occurrence matrix features, Hu moments, local binary patterns, histogram of oriented gradients, and color invariants as traditional features for injection. We tested the proposed method on the EuroSAT dataset to perform land classification. Our experimental results show that the proposed method significantly improves the land classification accuracy especially when applied to small-sized CNN models.
\end{abstract}


\section{Introduction}

Deep learning methods have proven themselves in recent years with their success in remote sensing applications \citep{Zhu, Ma, Liang, Lu}. Among these, CNN based models have high accuracy in solving the land classification problem using satellite or aerial images. On the other hand, traditional feature extraction methods were the only option before the deep learning era to solve such problems. Hence, they became mature in time. These two methods, deep learning based and traditional, have been used separately most of the times up to now.

Successful CNN models usually have large size due to the number of parameters they have. Therefore, they need significant processing power and memory space. Although this may not cause a problem in server-based or cloud applications, it limits the usage of such large sized models on low-power embedded platforms such as the ones on UAV \cite{Osco}. Several studies emphasize this limitation on object detection and land classification via UAV platforms \citep{Nuri, Wang, Rui}. The main bottleneck for these systems is the computing power limits. Hence, the model size, complexity, and data processing capacity of the hardware are of great importance, especially in real-time image processing tasks. Therefore, quantization, pruning, filter compression, and matrix factorization are usually applied to the CNN model at hand \cite{Goel}. The aim here is to decrease the complexity of the model such that it becomes suitable to be deployed on embedded low-powered devices. However, the model performance and accuracy of the model generally decreases as a result of these operations.

In this study, we propose a novel method to increase the performance of CNN models, especially the ones with small size. Our method injects traditional features to CNN models for this purpose. The method is developed especially for low-complexity models deployed on embedded systems such as the ones on UAVs. Detailed literature review reveals that such a method has not yet been used in remote sensing as we proposed. The closest study was developed by \cite{Jbene} in which statistical features are used to increase the performance of the CNN model. The authors did not report a significant performance increase with their method.

Next, we introduce the candidate CNN models used in this study. Then, we summarize the traditional feature extraction methods that we use in the proposed method. Afterward, we explain the feature injection method we proposed in detail. We next provide the performance of the proposed method on the EuroSAT dataset to for land classification. Finally, we provide comments and ways of extending the proposed method.

\section{Candidate CNN Models for Land Classification}

We picked five candidate CNN models to be used in the proposed method. We briefly summarize them in this section. While using these models, we apply transfer learning by freezing the top layer of each model with pre-trained "ImageNet" weights. ImageNet is the dataset used for classification tasks \citep{5206848}. We provide the number of parameters and model size for these models at the end of the section.

\subsection{SqueezeNet}

The first CNN model we picked in this study is SqueezeNet proposed by \cite{squeezeNet}. This model has a small size specifically developed to be used in embedded systems. Therefore, it represents the low end of the models available in literature. After freezing the top layer, we consider 512 SqueezeNet features (tensors) to be used in Section~\ref{section:feature_injection}.

\subsection{MobileNetV2}

MobileNetV2 is the next CNN model considered in this study. It has been proposed by \cite{mobilenetv2}. As in SqueezeNet, it has small model size such that it can be safely used on mobile and edge devices. After freezing the top layer, we consider 1280 MobileNetV2 features (tensors) to be used in Section~\ref{section:feature_injection}.

\subsection{ShuffleNetV2}

ShuffleNetV2 is the third CNN model used in this study. It has been proposed by \cite{ShuffleNet}. ShuffleNetV2 has similar number of parameters and model size as in MobileNetV2. Hence, it also represents the low model size option. After freezing the top layer, we consider 1024 ShuffleNetV2 features (tensors) to be used in Section~\ref{section:feature_injection}.

\subsection{VGG16}

Besides small sized models considered in previous sections, we also picked large models in this study. The aim for this selection is to observe the effect of the proposed feature injection method on them. Therefore, we picked the VGG16 model proposed by \cite{VGG16}. After freezing the top layer, we consider 512 VGG16 features (tensors) to be used in Section~\ref{section:feature_injection}.

\subsection{ResNet50V2}

ResNet50V2 is the fifth and final CNN model considered in this study. It has been proposed by \cite{ResNet50V2}. The reason for choosing this model was to see how the largest model that we have selected, would behave in the proposed method. After freezing the top layer, we consider 2048 ResNet50V2 features (tensors) to be used in Section~\ref{section:feature_injection}.

\subsection{Summary of the Candidate CNN Models}

We summarize the CNN models considered in the previous section next. The aim is to compare these models in terms of their number of parameters and mode size. Therefore, we tabulate these values in Table~\ref{tab:comparisonCNN}. As can be seen in this table, the selected CNN models have size between 0.5 MB and 528 MB. The number of parameters in these models are within the range of 729 K to 23.5 M. We will use these models while injecting traditional features to them.

\begin{table}[htbp]
\tbl{The parameter and model size of the candidate CNN models.}
{
\begin{tabular}{@{}ccc@{}}
\toprule
\textbf{Model}& \textbf{\# Parameters} & \textbf{Model Size} \\
\midrule
SqueezeNet & 729 K & 0.5 MB\\
MobileNetV2 & 2.2 M & 14 MB\\
ShuffleNetV2 & 4 M & 10 MB \\
VGG16 & 14.7 M & 528 MB \\
ResNet50V2 & 23.5 M & 98 MB\\
\bottomrule
\end{tabular}
}\label{tab:comparisonCNN}
\end{table}

\section{Traditional Features for Land Classification}

Our aim in this study is to improve the performance of known CNN models by injecting traditional features to them. Therefore, we summarize the traditional feature extraction methods in this section. To note here, the proposed system is not limited by the features extracted in this section. Other traditional feature extraction methods can also be used in the proposed framework.

\subsection{Sample Mean}

The first traditional feature extraction method used in this study is the sample mean. A mean value can be considered as a first-order statistical feature extraction method that can be used to summarize information in color bands. Therefore, we obtain the average (sample mean) normalized value for each band in the color image at hand. Normalization refers to re-scaling real-value numbers into a 0 to 1 range. As a result, we extract three features based on sample mean.

\subsection{Gray Level Co-occurrence Matrix Features}

Another traditional feature extraction method considered in this study is based on gray level co-occurrence matrix (GLCM) proposed by \cite{haralick} for texture analysis. This method has been generalized to other problems, including land use classification in satellite images. Haralick~\emph{et al.} introduced several second-order statistical features based on GLCM as correlation, contrast, homogeneity, energy, and ASM matrix features calculated on the grayscale image at hand. We use them as the second set of traditional features in this study.

\subsection{Hu Moments}

The third traditional feature extraction method used in this study is moments proposed by \cite{hu}. In this study, we calculate seven such features obtained from the grayscale image at hand.

\subsection{Local Binary Patterns}

The fourth traditional feature extraction method used in this study is the local binary patterns proposed by \cite{lbp}. This method has been initially developed for texture analysis. As in other methods, it has also been applied to land use classification studies. We extracted LBP 64 features from the grayscale image at hand.

\subsection{Histogram of Oriented Gradients}

The fifth traditional feature extraction method used in this study is the histogram of oriented gradients (HOG) proposed by \cite{hog}. It has been widely used in computer vision to solve object detection problems. HOG counts the occurrence of gradient orientations of a portion of the image. After using the grayscale data in the HOG method, we collected HOG values created for the single color band in a one-dimensional array, as we did in the LBP. As a result, we obtain 64 features for the image at hand.

\subsection{Color Invariants}

The sixth and final traditional feature extraction method used in this study is based on color information. We extract it via color invariants introduced by \cite{colorinvariance}. As a result, we obtain 64 features for the image at hand.

\section{Proposed Feature Injection Method for Land Classification}\label{section:feature_injection}

We can divide a CNN model into two parts as feature extraction and classification. The feature extraction part extracts features from a given source via successive filtering and nonlinear operations. Here, filter coefficients are learnt via training. Hence, an adaptive structure can be formed for the image set at hand. Thanks to the transfer learning method that we are using, we do not learn filter coefficients every time. Instead, we learn them for the ImageNet dataset once and keep these coefficients as fixed. The classification part is composed of a fully-connected neural network (FCNN) most of the times.

\begin{figure}[htbp]
	\centering
	\includegraphics[width=6cm]{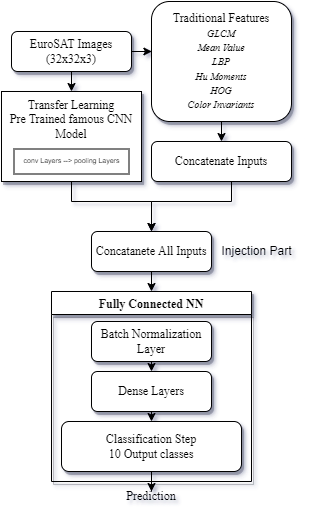}
	\caption{Visual representation of the proposed feature injection method.}
	\label{fig:wideDeep}
\end{figure}

We provide the visual representation of the proposed feature injection method in Figure~\ref{fig:wideDeep}. As can be seen in this figure, the CNN feature extraction part provides its own features. Traditional feature extraction methods extract their own features. Then, these two feature sets are merged with the concatenating method. Hence, the concatenate layer takes both inputs, and returns a single vector as merging of all inputs just before the classification step of the CNN model. This way, we inject the traditional features to the CNN model.

The concatenated traditional and CNN features are fed to the FCNN for the land use classification operation. We used max pooling and batch normalization operations just before concatenating the traditional features. We have 10 classes in our problem. Hence, the FCNN has 10 outputs. We train the parameters of the classification part (FCNN) of each CNN model for the problem at hand. While doing so, we also adjust weights of the injected traditional features fed to the FCNN. Hence, the overall model is reconfigured to solve the land classification problem at hand.

\section{Experiments}

We conducted experiments to measure the performance of the proposed method in this section. The body of the methods are written in Python language to run on a Jupyter notebook. The hardware power comes from the Google Colaboratory environment. We used Keras with the TensorFlow framework in implementation. During training, the epoch number is set to 16 with batch size 64. In optimizer side "Adam" is used in training. Each test has been repeated five times and the average values are listed in the following sections. The test code block can be accessed from Github by \cite{Aksoy_Object_Recognition_in}.

\subsection{Dataset used in Experiments}

We tested the proposed method on the EuroSAT dataset announced by \cite{eurosat} as a benchmark for land cover classification tasks. This dataset consists of 27000 RGB images with a resolution of 10  meters and hold labels with information of 10 different classes as annual crop, forest, herbaceous vegetation, highway, industrial, pasture, permanent crop, residential, river, and sea lake. We picked 80\% of the dataset for training and the rest is used for testing. Hence, we have 21600 and 5400 images for training and testing, respectively.

\subsection{The Effect of Injecting Traditional Features to the SqueezeNet Model}

In order to measure the performance of our proposed method, we first used the SqueezeNet model as a base. We injected different traditional feature sets to the model and tabulated the obtained results in Table~\ref{table:squezenet}. As can be seen in this table, the improvement in accuracy reaches 8.4\% when all traditional features are injected to the SqueezeNet model.

\begin{table}[htbp]
\tbl{The effect of injecting traditional features to the SqueezeNet model.}
{
\begin{tabular}{@{}lcc@{}}
\toprule
\textbf{Test Scenario} &  \textbf{Maximum Accuracy} \\
\midrule
SqueezeNet & 0.6778  \\
SqueezeNet + GLCM features& 0.6656  \\
SqueezeNet + color invariants & 0.7056\\
SqueezeNet + Hu moments + HoG + LBP + sample mean & 0.7403 \\
SqueezeNet + all traditional features & \textbf{0.7618} \\
\bottomrule
\end{tabular}
}
\label{table:squezenet}
\end{table}

\subsection{The Effect of Injecting Traditional Features to the MobileNetV2 Model}

We next used the MobileNetV2 model as the base to measure the performance of the proposed method. As in the previous section, we injected different traditional feature sets to the model and tabulated the obtained results in Table~\ref{table:mobilenet}. As can be seen in this table, the improvement in accuracy reaches 2.27\% when all the traditional features are injected to the MobileNetV2 model.

\begin{table}[htbp]
\tbl{The effect of injecting traditional features to the MobileNetV2 model.}
{
\begin{tabular}{@{}lcc@{}}
\toprule
\textbf{Test Scenario} & \textbf{Maximum Accuracy} \\
\midrule
MobileNetV2 & 0.6062  \\
MobileNetV2 + GLCM features & 0.6101  \\
MobileNetV2 + color invariants & 0.6160\\
MobileNetV2 + Hu moments + HoG + LBP + sample mean & 0.6208 \\
MobileNetV2 + all extracted features & \textbf{0.6289} \\
\bottomrule
\end{tabular}
}
\label{table:mobilenet}
\end{table}

\subsection{The Effect of Injecting Traditional Features to the ShuffleNetV2 Model}

Third, we used the ShuffleNetV2 model as the base to measure the performance of the proposed method. As in the previous section, we injected different traditional feature sets to the model and tabulated the obtained results in Table~\ref{table:shufflenet}. As can be seen in this table, the improvement in accuracy reaches 4.96\% when all traditional features are injected to the ShuffleNetV2 model.

\begin{table}[htbp]
\tbl{The effect of injecting traditional features to the ShuffleNetV2 model.}
{
\begin{tabular}{@{}lcc@{}}
\toprule
\textbf{Test Scenario} & \textbf{Maximum Accuracy} \\
\midrule
ShuffleNetV2 & 0.8502  \\
ShuffleNetV2 + GLCM features & 0.8623  \\
ShuffleNetV2 + color invariants & 0.8862\\
ShuffleNetV2 + Hu moments + HoG + LBP + sample mean & 0.8971 \\
ShuffleNetV2 + all extracted features & \textbf{0.8998} \\
\bottomrule
\end{tabular}
}
\label{table:shufflenet}
\end{table}

\subsection{The Effect of Injecting Traditional Features to the VGG16 Model}

Next, we used the VGG16 model as a base in order to measure the performance of our proposed method. As in the previous section, we injected different traditional feature sets to the model and tabulated the obtained results in Table~\ref{table:vgg16}. As can be seen in this table, the improvement in accuracy is only 0.0014\% when all traditional features are injected to the VGG16 model.

\begin{table}[htbp]
\tbl{The effect of injecting traditional features to the VGG16 model.}
{
\begin{tabular}{@{}lcc@{}}
\toprule
\textbf{Test Scenario} &  \textbf{Maximum Accuracy} \\
\midrule
VGG16 & 0.9091  \\
VGG16 + GLCM features& 0.8995  \\
VGG16 + color invariants & 0.9018\\
VGG16 + Hu moments + HoG + LBP + sample mean & 0.9003 \\
VGG16 + all extracted features & \textbf{0.9105} \\
\bottomrule
\end{tabular}
}
\label{table:vgg16}
\end{table}

\subsection{The Effect of Injecting Traditional Features to the ResNet50V2 Model}

Finally, we used the ResNet50V2 model as a base in order to measure the performance of our proposed method. As in the previous section, we injected different traditional feature sets to the model and tabulated the obtained results in Table~\ref{table:resnet}. As can be seen in this table, the improvement in accuracy is only 0.0158\% when all traditional features are injected to the ResNet50V2 model.

\begin{table}[htbp]
\tbl{The effect of injecting traditional features to the ResNet50V2 model.}
{
\begin{tabular}{@{}lcc@{}}
\toprule
\textbf{Test Scenario} & \textbf{Maximum Accuracy} \\ \midrule
ResNet50V2  & 0.6720  \\
ResNet50V2 + GLCM features & 0.6802  \\
ResNet50V2 + color invariants & 0.6826\\
ResNet50V2 + Hu moments + HoG + LBP + sample mean & 0.6804 \\
ResNet50V2 + All extracted Features & \textbf{0.6878} \\ \bottomrule
\end{tabular}
}
\label{table:resnet}
\end{table}

\subsection{Overview of the Injection Performance}

We can summarize all the experiments performed in the previous section as follows. When the CNN model size is small, injecting traditional features may increase the accuracy significantly as in the SqueezeNet, MobileNetV2, and ShuffleNetV2 models. Among these, improvement in the SqueezeNet model reaches up to 8.4\% which is a significant improvement. To note here, added parameters to the model (as injecting traditional features) have the size increase of 66 KB. This addition can be accepted when the corresponding improvement in accuracy is considered. On the other hand, the improvement in accuracy for large sized models such as VGG16 and ResNet50V2 is negligible. One reason for this result is that, these large models extract almost all information from the image due to their excessive parameters. Hence, injecting traditional features to such models increase accuracy marginally.

\begin{figure}[htbp]
	\centering
	\includegraphics[width=10cm]{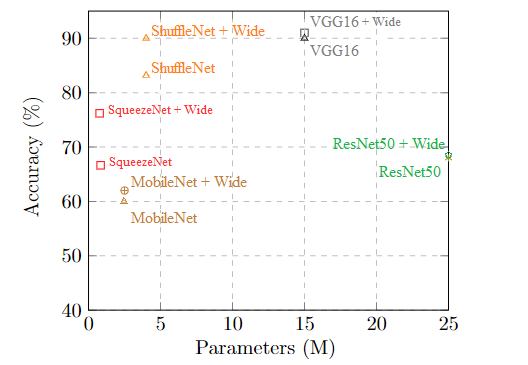}
	\caption{Model size vs. accuracy comparison of the selected CNN models before and after injecting traditional features.}
	\label{fig:paramvsacc}
\end{figure}

We summarized the accuracy improvements for the selected models in Figure~\ref{fig:paramvsacc}. This figure further justifies our observations such that the accuracy of relatively simple CNN models can be improved significantly by injecting traditional features to them.

\section{Final Comments}

In this study, we propose a novel method to improve the accuracy of CNN models, especially the ones with small size, by injecting traditional features to them. To test the effectiveness of the proposed method, we applied it to the CNN models SqueezeNet, MobileNetV2, ShuffleNetV2, VGG16, and ResNet50V2 having size 0.5 MB to 528 MB. We used the sample mean, gray level co-occurrence matrix features, Hu moments, local binary patterns, histogram of oriented gradients, and color invariants as traditional features for injection. We tested the proposed method on the EuroSAT dataset, consisting of 10 land use classes, to perform land classification. We observed that, the improvement in land classification accuracy reaches up to 8.4\% and 4.96\% when all traditional features are injected to the SqueezeNet and ShuffleNetV2 models, respectively. Hence, we can deduce that the proposed method significantly improves the land classification accuracy when applied to the small-sized CNN models. Unfortunately, such improvements in accuracy cannot be reached when the proposed method is applied to large-sized models. One reason for this result is that, these large models extract almost all information from the image due to their excessive parameters. Hence, injecting traditional features to such models increase accuracy only marginally. Increase in accuracy by the proposed method is achieved by adding only 350 extra traditional features. Hence, we can claim that the proposed method can be helpful in solving the land classification problem on edge (low power and low memory) devices.

\bibliographystyle{tfcad}
\bibliography{Injecting_traditional_to_deep}

\end{document}